\documentclass[10pt,twocolumn,letterpaper]{article}

\usepackage[pagenumbers]{cvpr}
\usepackage{color}
\usepackage{pifont}
\usepackage{amsmath}
\usepackage{afterpage}
\usepackage{cuted}
\usepackage{epigraph}
\usepackage[toc,page]{appendix}

\usepackage{tikz}
\usepackage{xcolor}

\definecolor{cvclgreen}{RGB}{0,100,0} 
\definecolor{clipblue}{RGB}{0,0,230}   

\def\linesolid{solid}

\newcommand{\inlinesymbol}[3][0.5]{
  \tikz[baseline={([yshift=-0.6ex]current bounding box.center)}]{
    \ifx#3\linesolid
      \draw[solid, color=#2, line width=1pt] (0,0) -- (#1,0);
    \else
      \draw[dashed, color=#2, line width=1pt] (0,0) -- (#1,0);
    \fi
    \filldraw[fill=#2, draw=#2] (#1/2,0) circle (0.07);
  }
}
\usepackage{natbib}
\usepackage{multirow}
\usepackage{graphicx}
\usepackage{makecell}
\usepackage{xcolor}
\usepackage[accsupp]{axessibility} 

\makeatletter
\renewcommand\paragraph{\@startsection{paragraph}{4}{\z@}
  {0.7ex \@plus 1ex \@minus .2ex} 
  {-1em}
  {\normalfont\normalsize\bfseries}}
\makeatother

\definecolor{cvprblue}{rgb}{0.21,0.49,0.74}
\usepackage[pagebackref,breaklinks,colorlinks,allcolors=cvprblue]{hyperref}

\title{Discovering Hidden Visual Concepts Beyond Linguistic Input in Infant Learning}

\author{
  Xueyi Ke\textsuperscript{\rm 1}\hspace{1.5em} 
  Satoshi Tsutsui\textsuperscript{\rm 1}\hspace{1.5em} 
  Yayun Zhang\textsuperscript{\rm 2}\hspace{1.5em} 
  Bihan Wen\textsuperscript{\rm 1}\thanks{Corresponding Author.} \\[0.5em]
  \textsuperscript{\rm 1}Nanyang Technological University\\
  \textsuperscript{\rm 2}The Max Planck Institute for Psycholinguistics
}

\begin{document}
\maketitle

\begin{strip}
\vspace{-1em}
\centering
\includegraphics[width=0.95\textwidth]{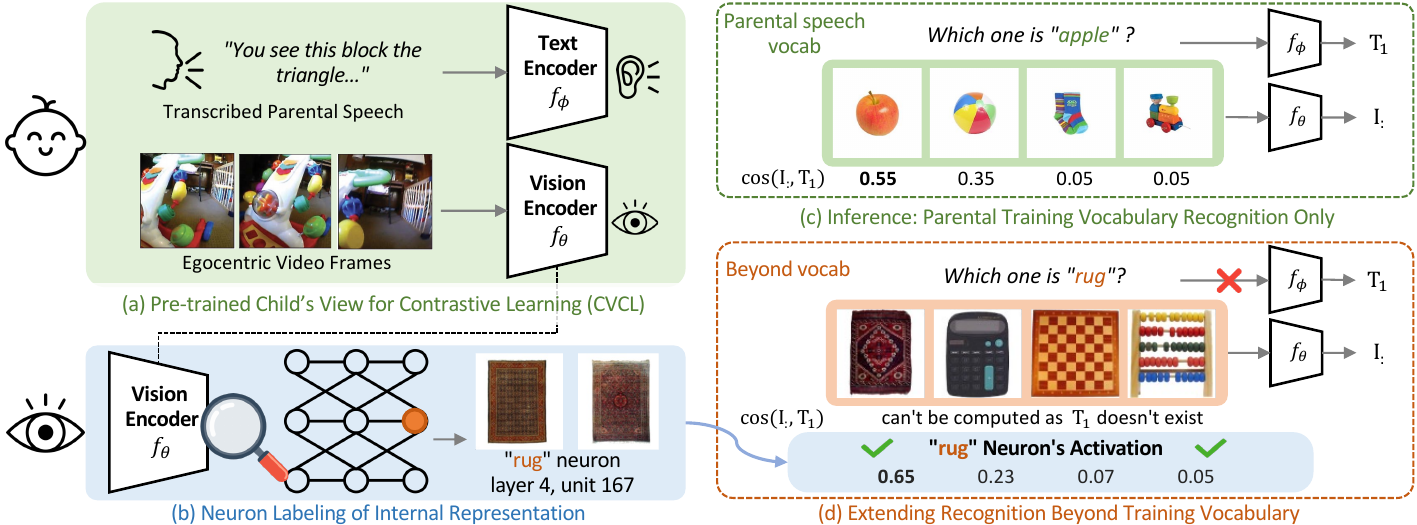} 
\captionof{figure}{
\footnotesize
    Inspired by the development of the infant visual system, CVCL~\cite{vong2024grounded} is trained on infant egocentric frames and transcribed parental speech \textcolor[rgb]{0.306, 0.478, 0.2}{\textbf{(a)}} and demonstrates object recognition ability within the vocabulary provided by parental speech \textcolor[rgb]{0.267, 0.447, 0.157}{\textbf{(c)}}. However, infant visual development is not limited to parental guidance. Thus, we hypothesize that a computational model trained on an infant’s daily experiences can similarly acquire visual concepts beyond its training parental speech. To explore this, we perform neuron labeling \textcolor[rgb]{0.145, 0.373, 0.655}{\textbf{(b)}} to identify visual concept neurons, including concepts that were never mentioned in the parental speech vocabulary (\eg, ``\textit{rug}'').. Based on discovered neurons, we show that the model can recognize objects beyond the training vocabulary, akin to early infant visual development \textcolor[rgb]{0.718, 0.271, 0.063}{\textbf{(d)}}. Some images are cited from prior work~\cite{vong2024grounded} as the original data is not accessible to us.
} \label{fig:teaser}
\end{strip}

\begin{abstract}
Infants develop complex visual understanding rapidly, even preceding of the acquisition of linguistic skills. As computer vision seeks to replicate the human vision system, understanding infant visual development may offer valuable insights. In this paper, we present an interdisciplinary study exploring this question: can a computational model that imitates the infant learning process develop broader visual concepts that extend beyond the vocabulary it has heard, similar to how infants naturally learn? To investigate this, we analyze a recently published model in Science by Vong et al., which is trained on longitudinal, egocentric images of a single child paired with transcribed parental speech. We perform neuron labeling to identify visual concept neurons hidden in the model’s internal representations. We then demonstrate that these neurons can recognize objects beyond the model’s original vocabulary. Furthermore, we compare the differences in representation between infant models and those in modern computer vision models, such as CLIP and ImageNet pre-trained model. Ultimately, our work bridges cognitive science and computer vision by analyzing the internal representations of a computational model trained on an infant visual and linguistic inputs. The project page is available at \url{https://kexueyi.github.io/webpage-discover-hidden-visual-concepts}.

\end{abstract}

\section{Introduction}\label{sec:intro}

Infants are remarkable learners, sparking interest across various academic disciplines. Computer vision is no exception, with researchers studying infant visual learning from various perspectives~\cite{bambach2018toddler, orhan2020self, sheybani2024curriculum, orhan2024learning, vong2024grounded}. 
A recent milestone is Child’s View for Contrastive Learning (CVCL) by Vong et al.\cite{vong2024grounded}, which trains a model from scratch on longitudinal egocentric videos of a single infant (6–25 months) paired with transcribed parental speech to learn visual-linguistic associations, similar to CLIP~\cite{radford2021learning}. The resulting model develops object recognition abilities, aligning with developmental psychology findings that infants acquire object names by linking words to visual referents~\cite{bergelson20126, golinkoff2013twenty}. In this paper, we analyze the internal visual representations of CVCL to better understand its mechanism, similar to how developmental studies observe infants' internal neurons~\cite{taga2003brain}. 

In developmental psychology, research on real infants suggests that their object recognition is deeply influenced by their visual experiences with the world. A headcam study of 8.5 to 10.5-month-old infants~\cite{clerkin2017real} showed that the first nouns infants acquire often correspond to objects they see most frequently. While infants may also hear the names of these frequently seen objects, their early visual familiarity plays a critical role in object recognition. Visual understanding may develop before the learning of corresponding names, potentially facilitating the process of word acquisition~\cite{phillips2000neurons}. For instance, newborn infants have shown an innate ability to recognize visual patterns~\cite{fantz1963balckwhitepattern}, and their development of visual concepts often precedes the emergence of verbal thought~\cite{mandler1992buildbaby}.

\textbf{Our Hypothesis}:  Based on these infant studies, we hypothesize that \textbf{
a computational model trained on an infant’s daily experiences may similarly acquire visual concepts extending beyond its linguistic training data.} While CVCL demonstrates visual-linguistic mappings of objects named in parental speech, its visual recognition capability should not be limited to parental vocabulary supervision. We conjecture that the vision encoder may develop the ability to recognize concepts beyond these linguistic training data, similar to how infants form object recognition before learning object names.

To investigate this, we analyze CVCL’s internal representations using network dissection~\cite{bau2017network, oikarinen2022clip}, or more intuitively, neuron labeling. Furthermore, we implement a neuron-based, training-free classification framework (\textit{NeuronClassifier}) that leverages visual concepts identified via neuron labeling. Relying solely on these neurons, this approach not only achieves better recognition performance than originally reported~\cite{vong2024grounded}, but also discovers internal visual concepts extending beyond the model's training vocabulary, supporting our aligned hypothesis.

From a developmental psychology perspective, the visual concepts discovered beyond the training vocabulary tend to have a higher age of acquisition (AoA)~\cite{kuperman2012aoa} values (Figure~\ref{fig:AoA}). This results supports CVCL's ability to develop visual understanding that precedes explicit labeling, mirroring cognitive studies where infants develop pre-verbal visual concepts~\cite{mandler1992buildbaby}. This likely reflects real-world learning dynamics: children first acquire labels for frequent, concrete concepts (captured in CVCL's training vocabulary), while visually grounded representations for unlabeled concepts still form earlier than their eventual linguistic acquisition. 

To further explore CVCL's internal representations in the context of computer vision, we compare its visual features with widely-used representations such as CLIP and ImageNet. While CVCL is trained on a unique infant dataset with limited exposure to diverse scenes and a much smaller dataset compared to CLIP, it exhibits similar low-level features in its early layers. However, its higher-level features in the final layer differ significantly. These differences also extend to the visual concept neurons in the model’s deeper layers.

\smallskip
\noindent \textbf{Contributions} The key contributions of our work are:

\begin{itemize}
    \item We show that infant model have developed understanding beyond linguistic training inputs, by discovering visual concepts hidden in the model representations, aligning with existing infant studies.
    \item We demonstrate that the discovered visual concept neurons can improve object recognition performance by implementing a training-free framework (\textit{NeuronClassifier}).
    \item We find that the infant model shares similar low-level representations with ImageNet or CLIP models but diverges in deeper layers due to a lack of diverse higher-level visual concepts.
\end{itemize}

\section{Related Work}\label{sec:related}
\paragraph{Learning from children.} 
Modeling how children learn has long been a strategy for advancing artificial intelligence.  Instead of directly replicating adult intelligence, Alan Turing suggested, ``why not rather try to produce one which simulates the child's?''~\cite{turing2009computing} Training models on egocentric videos or multimodal data captured from infant perspectives aligns with this idea.~\cite{bambach2018toddler, orhan2024learning, sheybani2024curriculum, orhan2020self,tsutsui2020computational, vong2024grounded}, as these videos approximate the input available to human infants during development. Our study established on CVCL~\cite{vong2024grounded}, which we explain in Section \ref{sec:cvcl-pre}.

\paragraph{Interpreting vision model representations.} Our goal is to understand the model internal neurons trained on infant data. In this context, techniques for interpreting intermediate representations in deep neural networks are relevant. Beginning with Network Dissection~\cite{bau2017network}, a method that quantifies alignment between hidden neurons and visual concepts, numerous studies have aimed to make black-box models more transparent. These methods enable compositional concept discovery~\cite{mu2020compositional}, assignment of compositional concepts with statistical quantification~\cite{bykov2024labeling}, open-vocabulary neuron captioning~\cite{hernandez2021natural}, and the use of CLIP’s rich embeddings for neuron-concept alignment~\cite{oikarinen2022clip, kalibhat2023identifying}. In addition to direct neuron dissection, other approaches analyze component functions by decomposing image representations to reveal the role of attention heads within multimodal embedding spaces~\cite{gandelsman2023interpreting} or identifying neurons with similar functions across a diverse model zoo~\cite{dravid2023rosetta}. 

\subsection{Preliminary: CVCL}\label{sec:cvcl-pre}

\paragraph{Training data.} CVCL is trained on the SAYCam-S dataset~\cite{sullivan2021saycam}, a longitudinal collection of egocentric recordings from a child aged 6 to 25 months, containing around 200 hours of video. To create meaningful image-text pairs for model training, transcripts were pre-processed to retain only child-directed utterances, excluding the child’s own vocalizations. Frames were extracted to align with utterance timestamps. The resulting dataset comprises 600,285 frames paired with 37,500 transcribed utterances, forming a multimodal dataset that simulates the sparse and noisy real-world experiences from which children learn.

\paragraph{Model architecture.} Employing a self-supervised contrastive learning approach akin to CLIP~\cite{radford2021learning}, CVCL learns to align egocentric visual frames with transcribed parental speech. Co-occurring pairs are treated as positive examples, while non-co-occurring pairs serve as negatives. This method allows the model to develop multimodal representations without external labels, imitating a child’s natural learning process.

\begin{figure}
    \centering
    \includegraphics[width=1\linewidth]{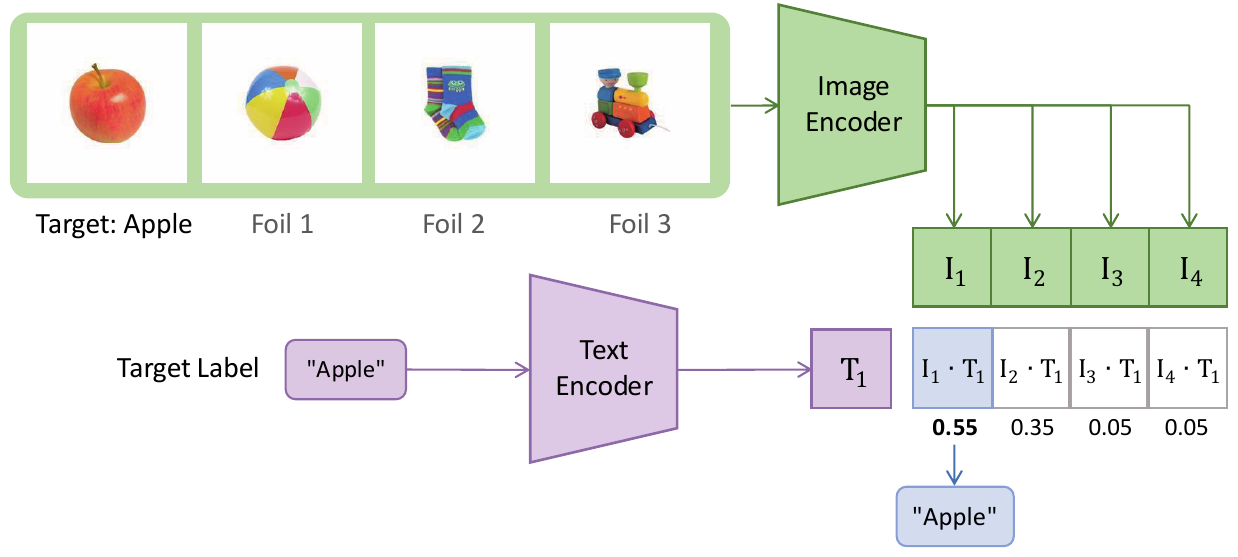}
    \caption{\textbf{CVCL's $n$-way evaluation}~\cite{vong2024grounded} poses a classification task of choosing the given target object label (\textit{``apple''}) from $n=4$ images, where only one of them contains the target object. The model feeds the target label into the text encoder and computes the pairwise similarities to each of the $n$ images, selecting the image with the highest similarity to demonstrate object recognition ability. However, this way of recognition limits its ability to the vocabularies in the text encoder. Our framework in Figure \ref{fig:NeuronClassifier} overcomes this limitation.}\label{fig:cvcl_eval}
\end{figure}

\paragraph{Evaluation.}\label{sec:cvcl's vanilla eval} For evaluation, CVCL adopts a $n$-way classification task (Figure~\ref{fig:cvcl_eval}) in which the model selects the most relevant visual reference from a set comprising one target image and $n-1$ foil images. This approach is inspired by the \textit{intermodal preferential looking paradigm (IPLP)}~\cite{golinkoff1987eyes, golinkoff2013twenty}, used in infant recognition studies to measure language comprehension through differential visual fixation. By aligning its visual and text encoders, CVCL achieves comparable in-domain test accuracy to models like CLIP. However, CVCL demonstrates relatively weak test performance on the Konkle object dataset~\cite{konkle2010conceptual}, which includes naturalistic object categories on a white background, using only classes available in the training data.

\section{Method}\label{sec:method}

In this section, we describe how to explore neuron-level concepts and leveraging them in $n$ way classification tasks. We begin by using published neuron labeling techniques to discover visual concepts hidden within CVCL~\cite{vong2024grounded}. Then, we introduce our framework to utilize these labeled neuron concepts for $n$-way classification.

\subsection{Neuron Labeling}

We follow CLIP-Dissect~\cite{oikarinen2022clip} for internal representation analysis due to its flexibility in concept sets and input image dataset.

\paragraph{Preliminary: CILP-dissect.}\label{sec:pre-clip} Given a neural network \( f(x) \), where \( f \) takes a  image \( x \) as input and \( x \in \mathcal{D}_{\textrm{probe}} \) with \( |\mathcal{D}_{\textrm{probe}}| = N \), and a concept set \( \mathcal{S} \) with \( |\mathcal{S}| = M \). The algorithm computes the concept-activation matrix \( P \in \mathbb{R}^{N \times M} \).
\begin{equation}
P_{i,j} = I_i \cdot T_j,
\end{equation}
where $I_i$ and $T_j$ are the embeddings of the images and concepts, respectively. For each neuron $k \in K$, where $K$ denotes the set of all neurons in the network, we summarize activations $A_k(x_i)$ with a scalar function $g$, producing an activation vector:
\begin{equation}
q_k = [g(A_k(x_1)), \ldots, g(A_k(x_N))]^\top \in \mathbb{R}^N.
\end{equation}
The neuron is labeled with the concept that maximizes similarity:
\begin{equation}
l_k = \arg\max_m \texttt  {sim}(t_m, q_k; P),
\end{equation}
where $\texttt{sim}(\cdot, \cdot)$ represents the similarity function (\eg cosine similarity or Soft-WPMI~\cite{wang2020towards, oikarinen2022clip}), and \textit{\( t_m\)} denotes the most similar text concept. Collecting the label $l_k$ for each neuron, we define the label vector  $\mathcal{L} = [l_1, l_2, \dots, l_K]$  to represent the assigned concepts across the entire model.

\begin{figure}[t!]
\centering
\includegraphics[width=1\linewidth]{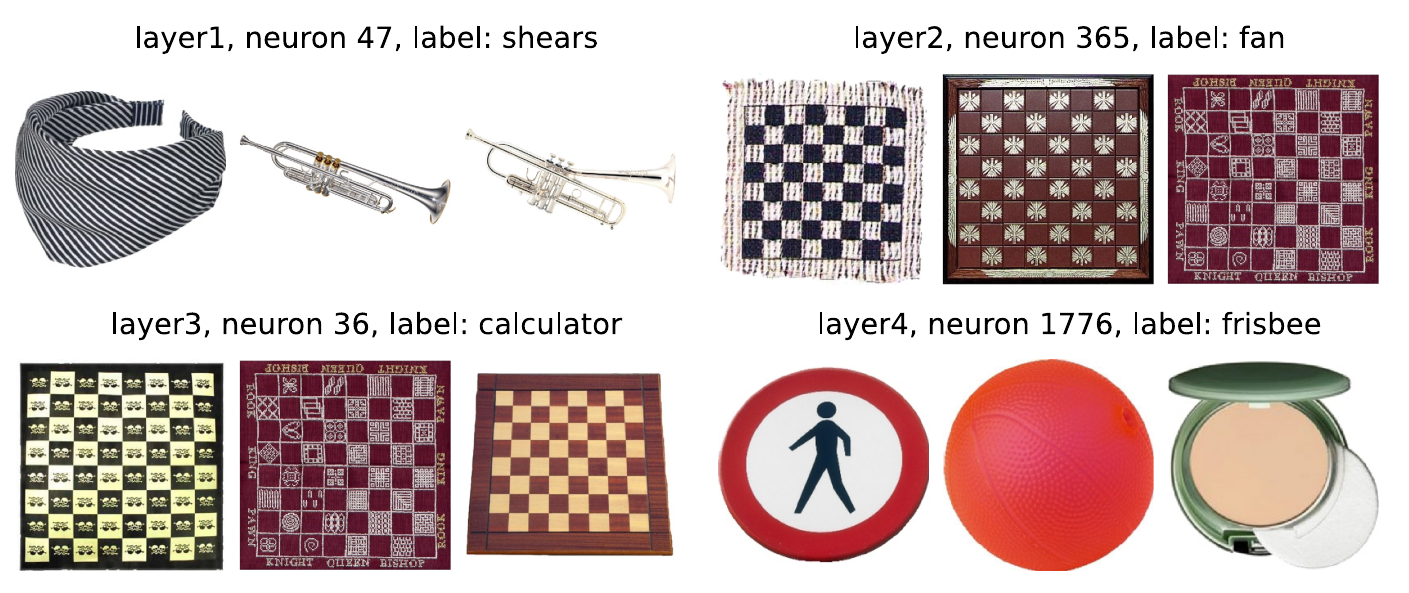}
\caption{\textbf{Spurious labeled neurons in infant model CVCL} by CLIP-Dissect, showing top-3 activated Konkle dataset images. This indicates: (1) not all neurons interpretable with human semantic concepts~\cite{elhage2022toy}; (2) neuron-labeling can produce spurious labels. To further support our hypothesis, we implement \textit{NeuronClassifier}.}
\label{fig:spurious_examples}
\end{figure}

During this neuron labeling process, we aim to assign meaningful concepts to each neuron. We use CLIP-dissect because:
\begin{itemize}
    \item \textbf{Concept set \(\mathcal{S}\):} 
    Instead of allowing an infinite range of possible concepts for neuron labeling, using a fixed concept set narrows this process by constraining it to a limited selection of concepts. Including diverse concepts in this set enables us to identify neurons corresponding to these visual concepts. This setup ensures that each neuron is assigned a specific label, though some labels may be spurious, as illustrated in Figure \ref{fig:spurious_examples}.
    
    \item \textbf{Probing dataset \(\mathcal{D}_{\textrm{probe}}\):} 
    The probing dataset allows neurons to activate specifically in response to the dataset of interest. For instance, when analyzing representations from an infant model, we use images from the Konkle object dataset, which better aligns with infant recognition than the ImageNet~\cite{deng2009imagenet} validation set. Additionally, for classification tasks, generalization can be achieved by adaptively selecting the probing dataset to focus on relevant concepts within a given dataset.
\end{itemize}

\paragraph{Extending beyond vocabulary.} 
The flexibility of the concept set $\mathcal{S}$ allowing us to discover visual concepts that the infant model has never encountered in its training data. In this process, CLIP serves as a well-pretrained miner, utilizing its rich image-text embeddings to identify concepts hidden within CVCL. By aligning the activations of CVCL’s neurons with CLIP’s embeddings, we discover meaningful hidden visual concepts within the infant model, even extending beyond the model’s linguistic training data. This approach leverages the diverse vocabulary of the concept set and the rich embeddings of CLIP to reveal visual concepts embedded in CVCL’s internal representations, each associated with corresponding neurons.

\subsection{Neuron-Based Classification}

\begin{figure}[ht]
    \centering
    \includegraphics[width=1\linewidth]{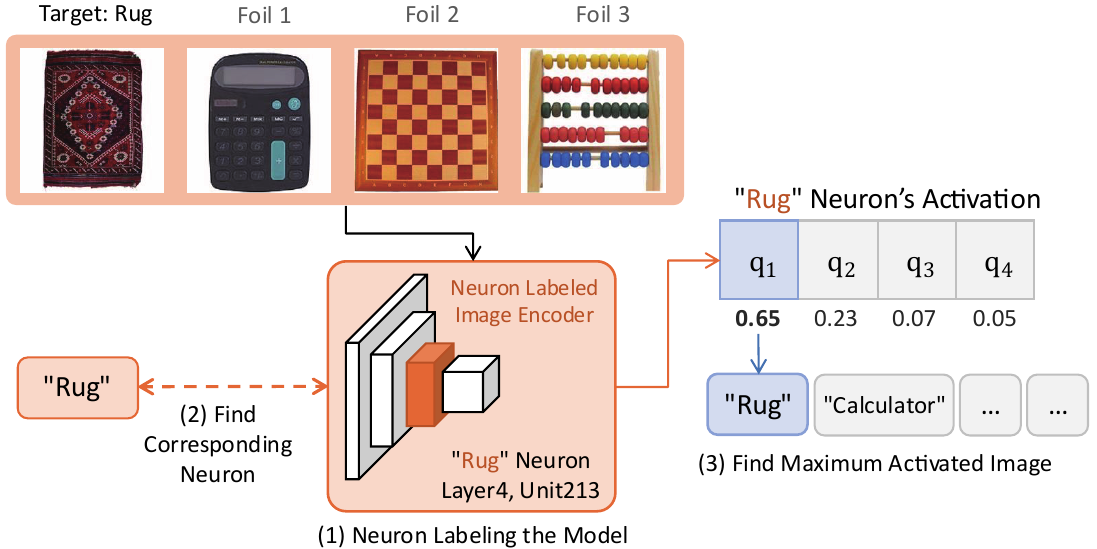}
    \caption{\textbf{Our \textit{NeuronClassifier} Overview.}
    A training-free $n$-way framework with three key steps: (1) Label all neurons in the network using a concept set that includes class labels and common words; (2) identify neurons associated with the target concept (\eg, \textit{``rug''}); (3) Evaluate the activations of visual concept neuron across $n$ images (in this example, $n=4$) and select the image with the highest activation as the most relevant to the target label.}
    \label{fig:NeuronClassifier}
\end{figure}

How do we ensure that neurons representing concepts beyond the model's original vocabulary truly exist within the network? In this section, we propose \textit{NeuronClassifier}, a training-free framework that leverages neuron activations to detect and validate such concepts. By discovering neurons with specific visual concept, we aim to confirm the presence of these latent, beyond-vocabulary neurons and use them to perform $n$-way classification. The framework, illustrated in Figure~\ref{fig:NeuronClassifier}, involves three main steps.

\paragraph{Step 1: neuron labeling with concept set.}
Given an image encoder $f(x)$, we labeled each neuron in the network using a concept set $\mathcal{S}$ that contains (but is not limited to) all class labels relevant to the task. The dissection process can be expressed as a function:
\begin{equation}
\mathcal{N}_s = \operatorname{Neuron Labeling}(f, \mathcal{S}),
\end{equation}
where $\mathcal{N}_s$ is the set of neurons labeled with concepts from $\mathcal{S}$. Each neuron $k \in \mathcal{N}_s$ is associated with a specific concept, such as \textit{``rug''} or \textit{``calculator''}, based on its alignment with concept embeddings obtained during neuron labeling (\eg similarity in CLIP-Dissect~\cite{oikarinen2022clip}).

\paragraph{Step 2: identifying visual concept neurons.}
Given a target label $l_k \in \mathcal{L}$, where $\mathcal{L}$ represents all neuron labels in the model, same as label vector in Section \ref{sec:pre-clip} (\eg, \textit{``rug''}), we select the subset of neurons labeled with this concept, denoted as $\mathcal{N}_{l_k}\subset \mathcal{N}_\mathcal{L}$. These neurons are responsible for encoding the target concept.

To further refine the labeling and reduce spurious assignments, we select the most similar neuron from $\mathcal{N}_{l_k}$ to represent the target concept. The similarity measure varies based on the dissection method used. For example, Network Dissection~\cite{bau2017network} employs Intersection over Union (IoU) for similarity, while CLIP-Dissect~\cite{oikarinen2022clip} supports multiple similarity metrics:
\begin{equation}
k^* = \arg\max_{k \in \mathcal{N}_\mathbf{L}} \texttt{sim}(t_{l_k}, q_k; P),
\end{equation}
where $t_{l_k}$ is the embedding of the target label $l_k$, $q_k$ is the neuron activation value, same in Section \ref{sec:pre-clip}. This step ensures that the neuron most aligned with the concept is selected, minimizing the possibility of spurious labeling.

For each selected neuron $k \in \mathcal{N}_{l_k}$, its activation value on an input image $x_i$ is computed as
\begin{equation}
q_k(x_i) = g\left(A_k(x_i) \right),
\end{equation}
where $A_k(x_i)$ is the raw activation map, and $g(\cdot)$ is a summary function (\eg, spatial mean) that reduces it to a scalar representing the neuron’s response strength.

\paragraph{Step 3: selecting the most relevant image.}
Given $n$ candidate images $\{x_1, x_2, \dots, x_n\}$ in an $n$-way trial, we compute the activation values $q_k(x_i)$ for neuron $k^*$ with highest similarity across all images. 
The image with the highest activation is selected as the most relevant to the target concept:
\begin{equation}
x^* = \arg\max_{x_i} q_{k^*}(x_i).
\end{equation}

In the example shown in Figure~\ref{fig:NeuronClassifier}, the target concept is \textit{``rug''}, and we select the image with the highest activation from the four candidates as the closest match to the concept.

\section{Neuron-wise Representation Analysis}\label{sec:micro}

In this section, we conduct a neuron-wise analysis of infant models and provide implementation details. First, we perform neuron labeling on the infant CVCL model. Then, we use our \textit{NeuronClassifier} framework to leverage visual concept neurons identified through neuron labeling, resulting in better recognition performance than the original approach~\cite{vong2024grounded} while also discovering internal visual concepts beyond the model's training vocabulary. These results support our hypothesis.

\subsection{Setup}

\paragraph{Datasets.} 
We use the Konkle object dataset~\cite{konkle2010conceptual}, as introduced in Section~\ref{sec:cvcl-pre}. This dataset consists of 3,406 images, each featuring a single object on a clean white background, including 406 test items across 200 classes. Each trial comprises $n$ images: one target image and the remaining as foils, with foil images randomly sampled from classes other than the target class. For each class, we generate 5 trials, each containing $n$ images. Following the previous work on CVCL, we use $n=4$ in our main experiments, with one target and three foils per trial.

\paragraph{Neuron labeling.} 
We utilize CLIP-Dissect~\cite{oikarinen2022clip} for neuron labeling, which assigns visual concepts to each neuron in the network. As we aim to perform classification on the Konkle object dataset, we use the same dataset as a part of $\mathcal{D}_{\textrm{probe}}$. Additionally, to avoid limiting the search space only around class names and ensure comprehensive neuron labeling, we employ a combined concept, consisting of three components:
\begin{itemize}
    \item \textbf{SAYCam-S vocabulary:} We clean the original vocabulary by removing noisy child speech and retaining meaningful words.
    \item \textbf{Common 
    English words:} We chose the top 30,000 most common English words based on a 1-gram frequency analysis by Peter Norvig~\cite{norvig_ngrams_beautiful, oikarinen2022clip}.
    \item \textbf{Class in Konkle object dataset:} All class labels from the Konkle object dataset are included.
\end{itemize}

We combine these three sources, ensuring no duplicates, resulting in a final concept set containing 30,427 words.

\paragraph{Models.} \label{sec:choice of models}
Our primary focus is the CVCL-ResNeXt50~\cite{vong2024grounded}, trained on SAY-Cam-S~\cite{sullivan2021saycam} dataset. For comprehensive analysis, we apply our framework to the following models:

\begin{itemize}
    \item \textbf{Infant models:} CVCL is trained on unique infant data, using both egocentric frames and transcribed parent speech. We also take DINO-S-ResNeXt50~\cite{orhan2024learning} as reference model compare recognition ability derived from same visual experience. It trained with DINO\cite{caron2021emerging} self-supervised approach on the infant same dataset.

    \item \textbf{Broadly-trained models:} To establish an upper bound, we include CLIP~\cite{radford2021learning} and ResNeXt50~\cite{xie2017aggregated}, which are broadly trained on large scale Internet images. Although CLIP uses a ResNet50-based vision encoder rather than ResNeXt, we select CLIP-ResNet50 due to the architectural similarity between ResNeXt~\cite{xie2017aggregated} and ResNet~\cite{he2016deep}.
    
    \item \textbf{Randomized model:} As a lower bound, we introduce a randomized version of the CVCL-ResNeXt50 model. In this setup, the convolution layer weights in the vision encoder are initialized using Kaiming Initialization~\cite{he2015delving}.
\end{itemize}

\subsection{Results}
In this section, we present our results and findings from applying neuron labeling and our \textit{NeuronClassifier} framework to the infant CVCL model, compared with other reference models. We define two class types for our analysis:

\begin{itemize}
\item \textbf{In-vocabulary classes}: Object classes present in the model's training linguistic input, which also appear in test object class and are detectable in internal representations.
\item \textbf{Out-of-vocabulary classes}: Object classes not included in the training linguistic input but detected in test object class and internal representations.
\end{itemize}

\begin{figure}[h]
    \centering
    \includegraphics[width=0.65\linewidth]{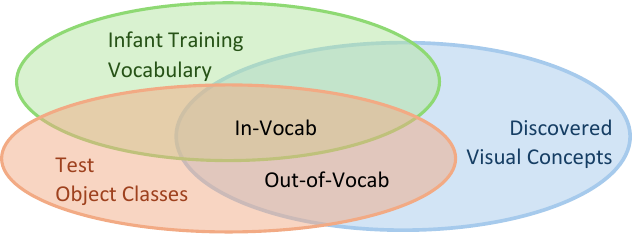}
    \caption{\textbf{In-vocabulary and out-of-vocabulary relationships} visualized using a Venn diagram.}
    \label{fig:venn_iv_oov}
\end{figure}

Our results demonstrate that the proposed framework effectively discovers meaningful neurons that represent concepts beyond the model's training vocabulary. These findings align with the cognitive perspective of vocabulary acquisition in infant development. We analyze the models' performance across different $n$-way classification settings, showing that our method yields strong out-of-vocabulary classification performance while simultaneously improving in-vocabulary classification accuracy.

\paragraph{Class coverage in visual concept neurons.}  

\begin{figure}[t]
    \centering
    \includegraphics[width=1\linewidth]{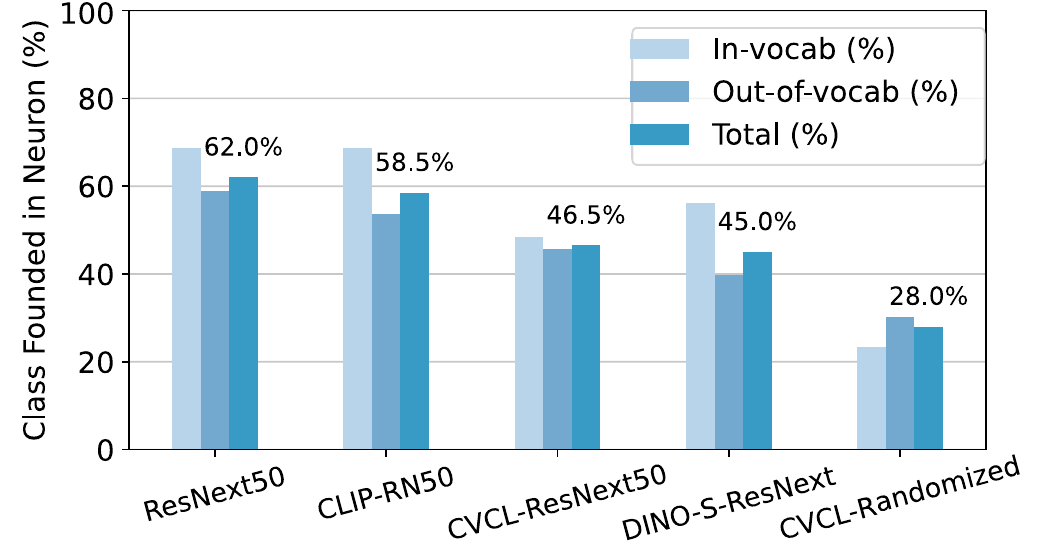} 

    \caption{\textbf{Class coverage in visual concept neurons.} Percentage of Konkle object dataset~\cite{konkle2010conceptual} classes identified through neuron labeling. Broadly pre-trained models (CLIP, ResNeXt) achieve over 50\% coverage, while developmentally inspired infant models (CVCL~\cite{vong2024grounded}, DINO~\cite{caron2021emerging}) show lower coverage. CVCL-Randomized provides a lower-bound comparison.}
    \label{fig:class_coverage}
\end{figure}

How well do visual concept neurons identified through neuron labeling correspond to specific class names (\eg, ``\textit{rug}'') rather than general descriptive attributes (\eg, ``\textit{red}'')? Figure~\ref{fig:class_coverage} shows the percentage of classes in the Konkle object dataset that are discovered in visual concept neurons from each model during the neuron labeling process. The results indicate that well-pretrained models, such as CLIP-ResNet50 and ResNeXt50, demonstrate broader class name coverage. While CVCL performs slightly weaker, it still maintains coverage of slightly less than 50\%. In contrast, CVCL-Randomized achieves only around 28\% coverage. This class coverage metric reflects the models' capacity to form class-corresponding meaningful representations during the neuron labeling process.

\paragraph{Age of Acquisition (AoA) ratings.}

Age of acquisition (AoA) is used to indicate when, and in what sequence, words are learned, and it is often assessed through ratings or observations reported by adults. This indirect method generally correlates well with other metrics indicating when children acquire vocabulary. Previous developmental work has shown that infants' early visual familiarity with common objects helps with object recognition, which subsequently helps support the process of learning the names of those objects. We next examined how early words in our models are learned and whether there is an AoA difference between in-vocab and out-of-vocab words. 

We used AoA ratings from a dataset compiled by Kuperman, Stadthagen-Gonzalez, and Brysbaert ~\cite{kuperman2012age}, which includes norms for over 30,000 English words gathered via Amazon Mechanical Turk. Each participant estimated the age in years at which they believed they first understood each word, even if they did not actively use them. This dataset is comparable to previously reported AoA norms~\cite{stadthagen2006bristol} gathered in laboratory settings.

\begin{figure}[t]
    \centering
    \includegraphics[width=1\linewidth]{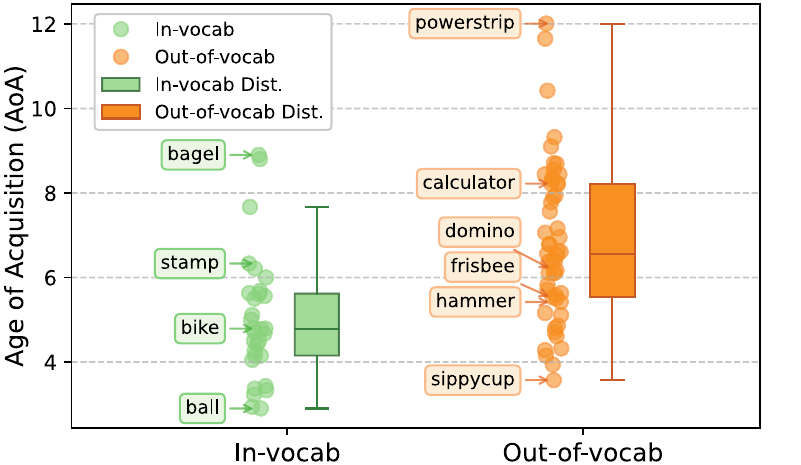}
    \caption{\textbf{Age of acquisition (AoA) ratings for in- and out-of-vocabulary visual concepts.} Comparison of word acquisition timing~\cite{kuperman2012age} between in-vocabulary and out-of-vocabulary concepts. Out-of-vocabulary concepts tend to have a higher estimated acquisition ages in the infant-inspired CVCL model, indicating development of visual understanding beyond explicit linguist inputs.}
    \label{fig:AoA}
\end{figure}

Using this set of AoA norms, we compared mean AoA between in-vocab and out-of-vocab words discovered in CVCL's internal representations. As shown in Figure~\ref{fig:AoA}, we found a significant difference between in-vocab and out-of-vocab AoA rating ( \textit{t}(82) = 4.64, \textit{p} $<$ 0.0001), in-vocab words (mean AoA = 4.99) are learned earlier than out-of-vocab words (mean AoA = 6.82). This pattern suggests that: (1) both sets of words are learned quite early, around later preschool and school years, with or without supervised labeling; (2) the difference in AoA between in-vocabulary and out-of-vocabulary words indicates that the infant model has developed a basic visual understanding of concepts with higher AoA. This foundational knowledge may lay the groundwork for word learning once corresponding parental speech is introduced.

\paragraph{Neuron-based classification performance.}

We evaluated the hidden potential of infant models' vision encoders by applying our \textit{NeuronClassifier} framework, summarized in Table~\ref{tab:in_out_vocab_results}. Despite being trained on infant egocentric data with limited amount and diversity, CVCL demonstrates the ability to recognize similarly as nature of infant learning, revealing strong out-of-vocabulary classification performance. This result suggests that this infant model developed broader visual concepts that extend beyond linguistic input, similar to how infants naturally learn. We also applied our method to in-vocabulary classification, where it outperformed the vanilla method previously used in CVCL~\cite{vong2024grounded}, as introduced in Figure ~\ref{fig:cvcl_eval}. ``All'' representing the combined performance on both ``in-vocab'' and ``out-of-vocab''. These results support our hypothesis. 

We include additional model comparisons:
(1) The DINO-S-ResNeXt50 infant model~\cite{orhan2024learning}, trained on the same dataset without text supervision, achieves comparable performance in visual representations. This suggests that models trained on identical data distributions with different self-supervised methods may yield similar representational outcomes.
(2) Broadly-trained models such as ResNeXt50 and CLIP establish performance upper bounds. However, CLIP underperforms in the vanilla setting, relying exclusively on visual neurons without text encoder guidance. While this work does not aim to advance zero-shot learning methods, it reveals visual concepts in infant models that emerge independently of linguistic inputs.

In classification on the Konkle object dataset, both DINO-S and ImageNet ResNeXt50 required fine-tuning for this task. Our framework enables neuron-based classification without downstream fine-tuning, and providing a training-free qualitative inspection of internal representations. 

\begin{table}[t]
    \caption{\textbf{Neuron-based classification results in $4$-way evaluation} among models in Section~\ref{sec:choice of models}.  
    ``\textit{Vanilla}'' refers to classification based on image-text similarity (Figure \ref{fig:cvcl_eval}). ``{\textcolor{red}{\ding{55}}}'' denotes cases where direct classification on the Konkle dataset~\cite{konkle2010conceptual} is not possible due to missing text encoder or need fine-tuning. By leveraging neurons discovered in the representation, \textit{NeuronClassifier} enables broader recognition, particularly in CVCL (bolded for emphasis), achieving improved recognition in both in-vocabulary and out-of-vocabulary, supporting our hypothesis. ``All'' represents the combined performance on both in- and out-of-vocabulary.
    }
\label{tab:in_out_vocab_results} 

\resizebox{\linewidth}{!}{
\centering

\begin{tabular}{c @{\hskip 1.5em} c @{\hskip 1em} c @{\hskip 0.5em} c @{\hskip 0.5em} c}
\toprule
\textbf{Method} & \textbf{Model} & \textbf{In-vocab} & \textbf{Out-of-vocab} & \textbf{All} \\
\midrule
\multirow{4}{*}{\textit{Vanilla}} 
& CLIP-ResNet50 & $98.81_{\textsubscript{$\pm$0.16}}$ & $96.93_{\textsubscript{$\pm$0.06}}$ & $97.42_{\textsubscript{$\pm$0.05}}$ \\
& ResNeXt50 & {\textcolor{red}{\ding{55}}} & {\textcolor{red}{\ding{55}}} & {\textcolor{red}{\ding{55}}} \\
& \makecell{\textbf{CVCL-ResNeXt50} 
}  & $\mathbf{36.18}_{\textsubscript{$\pm$0.91}}$ & {\textcolor{red}{\ding{55}}} & {\textcolor{red}{\ding{55}}} \\ 
& DINO-S-ResNeXt50 & {\textcolor{red}{\ding{55}}} & {\textcolor{red}{\ding{55}}} & {\textcolor{red}{\ding{55}}} \\
\midrule
\multirow{4}{*}{\makecell{\textit{Neuron} \\ \textit{Classifier}}}
& CLIP-ResNet50 & $91.59_{\textsubscript{$\pm$0.52}}$ & $88.66_{\textsubscript{$\pm$0.35}}$ & $89.79_{\textsubscript{$\pm$0.38}}$ \\
& ResNeXt50 & $88.17_{\textsubscript{$\pm$0.45}}$ & $93.28_{\textsubscript{$\pm$0.36}}$ & $91.88_{\textsubscript{$\pm$0.15}}$ \\
& \textbf{CVCL-ResNeXt50} & $\mathbf{79.50}_{\textsubscript{$\pm$0.78}}$ & $\mathbf{76.81}_{\textsubscript{$\pm$0.35}}$ & $\mathbf{77.79}_{\textsubscript{$\pm$0.40}}$ \\
& DINO-S-ResNeXt50 & $77.53_{\textsubscript{$\pm$0.24}}$ & $77.96_{\textsubscript{$\pm$0.27}}$ & $77.65_{\textsubscript{$\pm$0.21}}$ \\
\bottomrule
\end{tabular}
}
\end{table}

\paragraph{Analysis across $n$-way settings.}

\begin{figure}[t]
    \centering
    \includegraphics[width=1\linewidth]{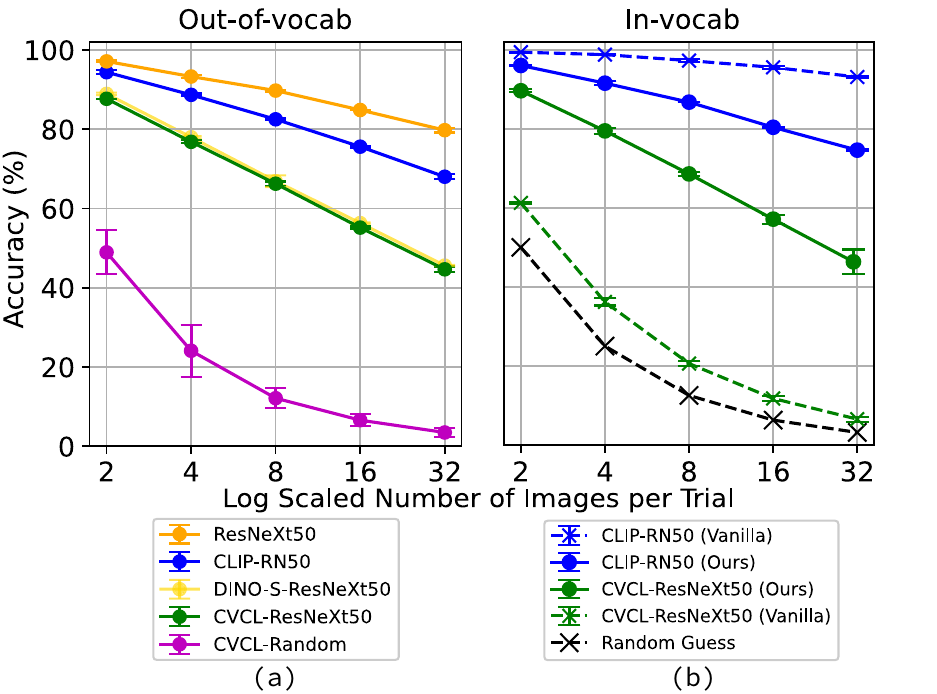}
    \caption{\textbf{In- and out-of-vocabulary class performance across $n$-way settings} using \textit{NeuronClassifier}. \textbf{(a)} For out-of-vocabulary classification (left), our method enables classification without additional training. The infant model CVCL
    (\inlinesymbol[0.3]{cvclgreen}{\linesolid})
    maintains robust performance as $n$ increases. \textbf{(b)} For in-vocabulary classification (right), ``ours'' (\textit{NeuronClassifier}) enhances CVCL recognition ability compared to the ``vanilla'' setting. However, CLIP
    (\inlinesymbol[0.3]{clipblue}{\linesolid})
    declines, as it relies solely on neurons without text encoder input. 
    Random ones serve as lower bounds.
    }
    \label{fig:in-out-of-vocab-nway}
\end{figure}

We evaluate the models under various $n$-way classification setups. Figure~\ref{fig:in-out-of-vocab-nway} illustrates the performance trends for in- and out-of-vocabulary class classification accuracy applying our \textit{NeuronClassifier}. 

Our method not only enables out-of-vocabulary classification but also significantly improves in-vocab performance compared to previous results~\cite{vong2024grounded}, further supporting the presence of beyond-vocabulary potential in infant models. However, due to limited class coverage, CVCL with our method can classify a maximum of 31 classes (see Appendix \ref{sec:n-way_31_clarification}). These findings support our hypothesis that the infant model has acquired visual concepts beyond its initial vocabulary. These results show that leveraging the model's internal representations for classification that go beyond its vocabulary is not only feasible but also robust across different $n$-way settings. 
These findings support our hypothesis that the infant model has acquired visual concepts beyond its initial vocabulary.

However, as $n$ increases exponentially, performance gradually declines. This decline is reasonable, as the task becomes increasingly difficult by nature as $n$ increases. It may also be attributed to dimensionality reduction in activation maps, leading to coarser classification. CLIP-RN50 and ResNeXt50 perform well with \textit{NeuronClassifier}, though not as effectively as their direct or fine-tuned versions, since our method is designed to reveal latent concepts rather than to perform fully optimized classification. In in-vocab settings, the ``vanilla'' approach represents direct classification as shown in Figure~\ref{fig:cvcl_eval}.

\section{Layer-wise Representation Analysis}

How does the representation learned from infant data differ from other representations widely used in the computer vision community?  To explore this, we perform a layer-wise representation analysis using Centered Kernel Alignment (CKA)~\cite{kornblith2019similarity}. We compute the similarity between the representations of the infant model(CVCL), ImageNet-pretrained, and CLIP. Additionally, we apply neuron labeling techniques from a layer-wise perspective to identify unique visual concepts discovered at each layer between ImageNet and infant models.

\begin{figure}[t]
    \centering
    \includegraphics[width=1\linewidth]{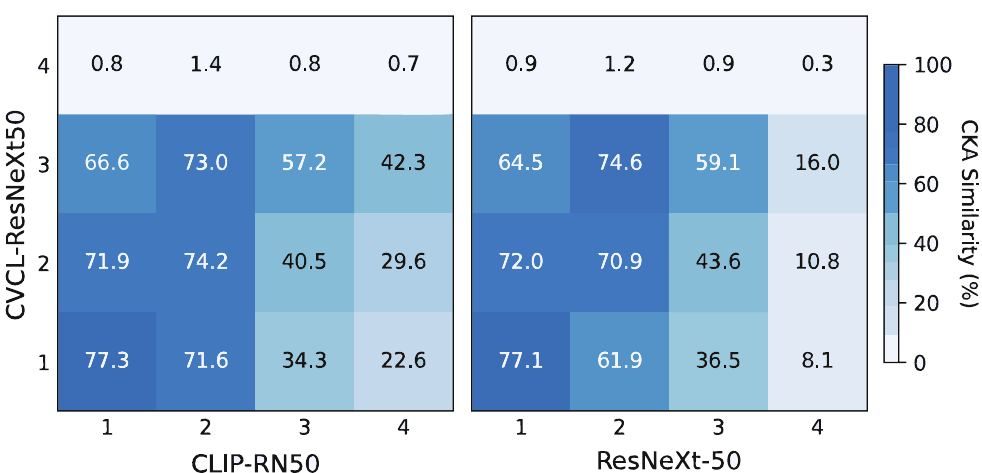}

    \caption{\textbf{CKA layer-wise similarity between CVCL and common models.} Using the ImageNet validation set as input, CVCL (y-axis) exhibits similarity to CLIP-RN50 (x-axis, left) and ResNeXt50 (x-axis, right) in the shallow layers (lower-level features) but diverges significantly in the final layer (higher-level features). Notably, Layer 4 of CVCL shows very low similarity to all layers of both common models.}
    \label{fig:cka_matrics}
\end{figure}

\paragraph{Layer similarity analysis.} 
CKA~\cite{kornblith2019similarity} applies HSIC~\cite{gretton2005kernel}
(see Appendix~\ref{sec:cka-defined}) 
over a set of images to provide layer-wise similarity scores between the representations of different neural networks. A higher CKA score indicates more similar representations between two models at the given layer. We use the ImageNet validation set~\cite{deng2009imagenet} as input for the networks, and compute the CKA similarity between the infant model (CVCL), the ImageNet-pretrained ResNeXt50, and CLIP-ResNet50. The results are presented as matrices in Figure~\ref{fig:cka_matrics}. The lower layers of CVCL exhibit greater similarity to larger-scale models than its deeper layers. In larger-scale models, shallow layers are known to capture low-complexity features, while deeper layers progressively specialize in capturing higher-level concepts~\cite{fel2024understanding, chen2023layer}. Therefore, CVCL -- the model trained on infant data -- successfully develops lower-level representations comparable to those in common pre-trained models. However, the divergence in deeper layers suggests a lack of the diverse higher-level representations typically observed in models trained on common datasets.

\paragraph{Neuron-based analysis.}

\begin{figure}[t]
    \centering
    \includegraphics[width=1\linewidth]{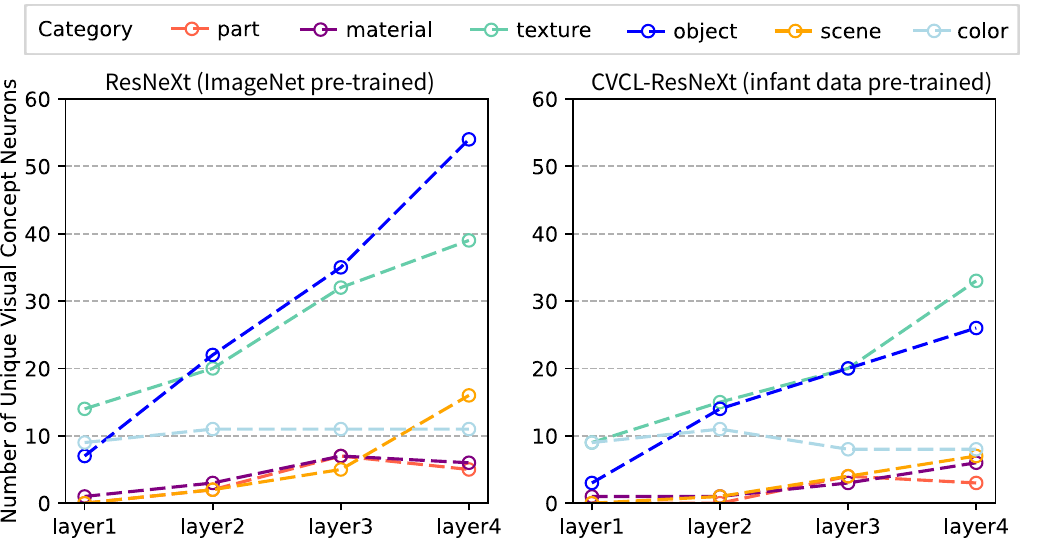}
    \caption{\textbf{Number of visual concepts in ImageNet and infant models.} Concepts are categorized using Broden dataset~\cite{bau2017network}. Neurons in deeper model layers capture increasingly complex concepts. Early layers primarily detect lower-level features like color and texture, while higher-level concepts such as objects and scenes emerge in deeper layers. CVCL exhibits fewer visual concepts than the ImageNet model, especially for higher-level visual concepts (\eg, objects and scenes).}
    \label{fig:net_unique}
\end{figure}

To investigate the characteristics of each layer, we apply network dissection to identify neurons that are aligned with specific visual concepts. Essentially, this is an extension of the neuron labeling process, where the Broden~\cite{bau2017network} dataset provides category labels for each visual concept neuron. We count the number of unique visual concepts discovered in each category and perform layer-wise comparisons to gain a broader view of the differences between models trained from ImageNet and infant data. In Figure~\ref{fig:net_unique}, we visualize the number of unique visual concept neurons across layers for each model. The results show that early layers in both models predominantly have neurons of low-level features like color, with minimal differences between models. However, as we move to deeper layers, higher-level concepts such as objects and scenes become more prominent, and the disparities between models become clearer. CVCL exhibits fewer unique visual concepts in these higher-level categories compared to ImageNet model. This finding aligns with the layer similarity analysis. 

\section{Conclusion}\label{sec:experiments}
In this paper, we explored whether an infant model (CVCL), trained on infant egocentric video frames and linguistic inputs, can acquire broader visual concepts extending beyond its initial training vocabulary. By introducing \textit{NeuronClassifier}, a training-free framework to discover and leverage visual concepts hidden in representations, we unlocked the CVCL visual encoder’s ability to recognize out-of-vocabulary concepts, establishing its potential as a strong classifier. Our findings also reveal that while CVCL, trained on a unique infant dataset with limited exposure to diverse scenes, it representations capture low-level features similar to those in common pre-trained models, they diverge significantly in higher-level representations, contributing to the observed performance differences.

Overall, our approach bridges cognitive science and computer vision, providing insights into how infant models develop visual concepts that precede linguistic inputs, aligning with the natural way infants explore the world through sight.

\paragraph{Limitations and future work.} 

Our study did not analyze the infant training data, as we were unable to access it due to limited access controls (see Appendix~\ref{sec:callopendataset}). Instead, we analyze models trained on infant egocentric data, finding developmental alignment with cognitive studies. However, we did not extend this analysis to adult egocentric data. While the study focuses on infant data for developmental process, the framework can be applied to adult models, which remains a direction for future research.

\clearpage
\section*{Acknowledgements} This work was partially supported by the National Research Foundation Singapore Competitive Research Program (award number CRP29-2022-0003). We thank Wai Keen Vong for invaluable discussion and CVCL's pretrained weights. 
We thank Jingyi Lin for figure discussions.
We thank the anonymous reviewers for their constructive feedback.

{
    \small
    \bibliographystyle{ieeenat_fullname}
    \bibliography{references}
}

\clearpage
\setcounter{page}{1}
\maketitle

\begin{appendices}

\section{Neuron-wise Analysis}
We present additional examples illustrating how the infant model perform classification using visual concept neurons. Furthermore, we provide complete results of Age of Acquisition (AoA) to quantify the cognitive level of visual concepts for both in-vocabulary and out-of-vocabulary cases.

\subsection{Neuron-based Classification Examples}
\begin{figure}[ht]
    \centering
    \includegraphics[width=1\linewidth]{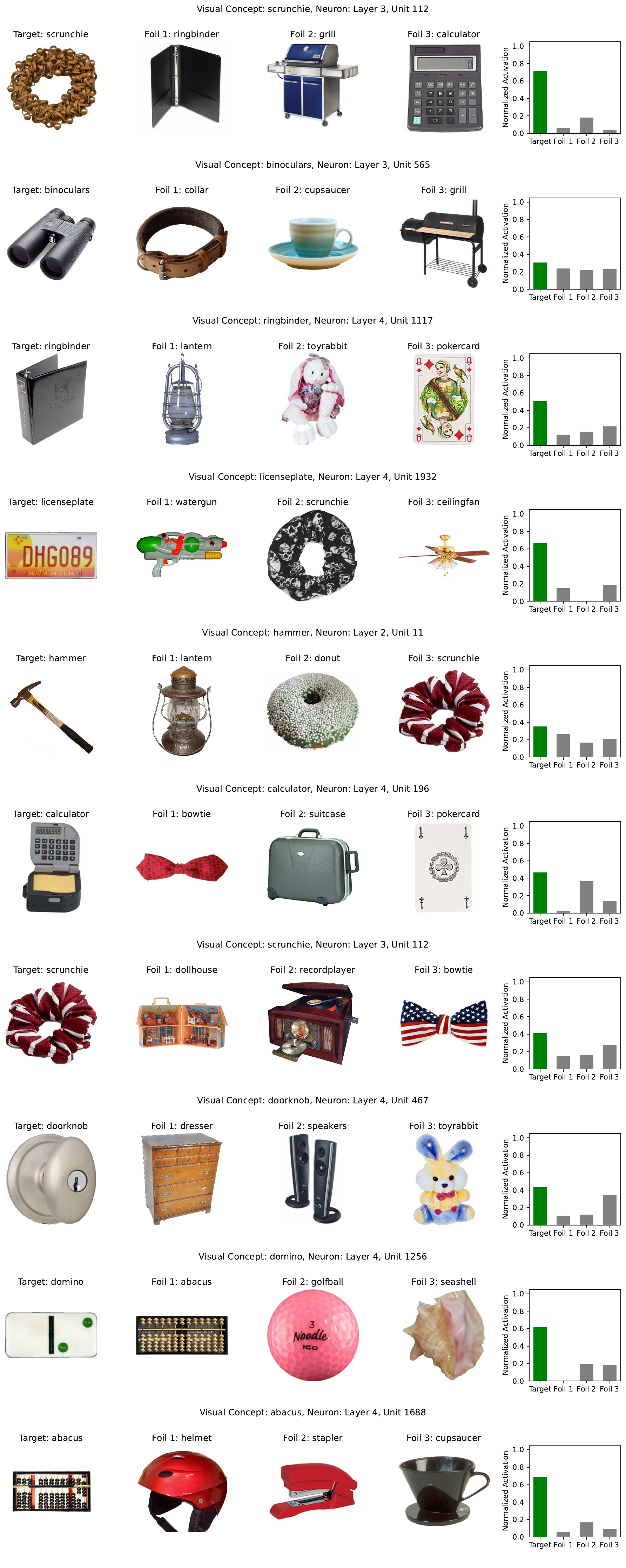}
\caption{\textbf{Correctly Classified Examples.} Green bars indicate the highest normalized activation values, corresponding to the target image for correct classifications. Subtitles display information about visual concept neurons. These examples represent out-of-vocabulary classes from the Konkle object dataset~\cite{konkle2010conceptual}.}
    \label{fig:random_correct_example}
\end{figure}

\begin{figure}[ht]
    \centering
    \includegraphics[width=1\linewidth]{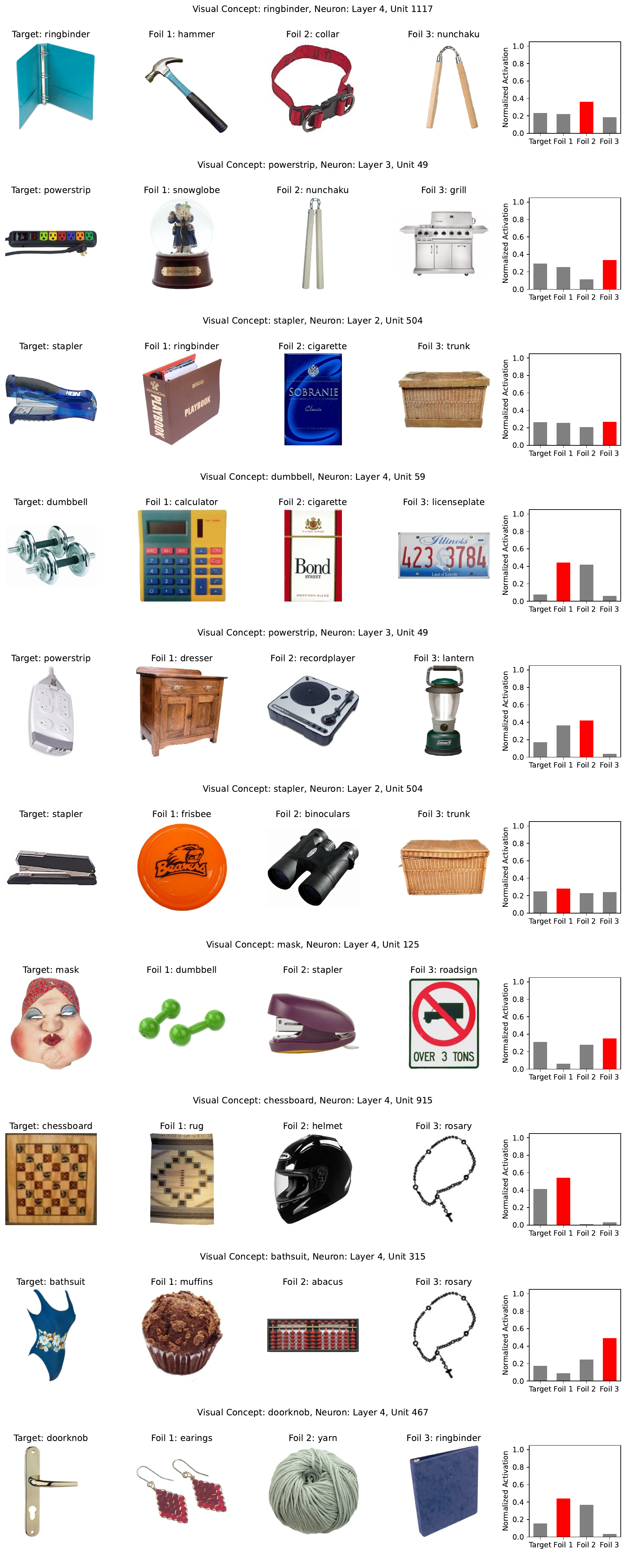}
\caption{\textbf{Incorrectly Classified Examples.} Red bars indicate the highest normalized activation values, corresponding to incorrect classifications. Subtitles display information about visual concept neurons. These examples represent out-of-vocabulary classes from the Konkle object dataset ~\cite{konkle2010conceptual}.}
    \label{fig:random_incorrect_example}
\end{figure}

For this analysis, we conducted 4-way classification experiments on the Konkle object dataset to evaluate out-of-vocabulary classification. The examples are derived from neurons selected randomly under the specified experimental settings, with classification trial images and neuron activations presented, details in Figure \ref{fig:random_correct_example} and Figure \ref{fig:random_incorrect_example}.

\subsection{Age-of-Acquisition Ratings}
Age-of-Acquisition (AoA) ratings, defined by Kuperman et al.~\cite{kuperman2012aoa}, estimate the age at which a person learns a word. These ratings were obtained via crowdsourcing using 30,121 English content words, organized into frequency-matched lists based on the SUBTLEX-US corpus~\cite{brysbaert2009moving}. Each list included calibrator and control words for validation. It strongly correlated with prior norms ($r = 0.93$ with Cortese and Khanna~\cite{cortese2008age}, $r = 0.86$ with the Bristol norms~\cite{stadthagen2006bristol}), confirming their reliability for studying vocabulary development.

Participants on Amazon Mechanical Turk rated the age they first understood each word on a numerical scale (in years). Words unfamiliar to participants could be marked with ``x'' to exclude outliers. Data cleaning removed non-numeric responses, ratings exceeding participant age, low-correlating responses ($r < 0.4$), and extremely high AoA ratings ($>25$ years). This yielded 696,048 valid ratings. 

\subsubsection{Detailed AoA Results}
Here we present visual concepts that from Konkle object dataset\cite{konkle2010conceptual} class, by applying neuron labeling, we found many visual concept neurons with corresponding class inside vision encoder's hidden representation. For founded classes, we investigate their AoA values to prove the alignment between computational model and infant cognition.

\begin{table}[h]
\label{iv-aoa}
\centering
\caption{\textbf{In-vocabulary Classes and Corresponding AoA Values.} The table lists the identified in-vocabulary classes along with their Age-of-Acquisition (AoA) values. For some classes, closely related words (shown in parentheses) were used to derive AoA values.}
\label{tab:iv-aoa} 
\resizebox{\linewidth}{!}{
\begin{tabular}{llll}
\toprule
Vocab (Col 1) & AoA (Col 1) & Vocab (Col 2) & AoA (Col 2) \\
\midrule
bike          & 2.9         &    abagel    & 4.79      \\
stamp         & 2.94        &    umbrell       & 4.79      \\
microwave     & 3.23      &    desk        & 5.00        \\
pen           & 3.33        &   hat         & 5.11    \\
knife         & 3.37        &     cookie      & 5.50         \\
broom         & 3.43        &    stool       & 5.56   \\
scissors      & 4.05        &  necklace    & 5.61         \\
  button    & 4.15        &    sofa        & 5.63     \\
   hairbrush     & 4.15        &      fan         & 5.68    \\
pizza         & 4.26        &    chair       & 6.00      \\
kayak         & 4.42        &     ball        & 6.21      \\
bucket        & 4.5         &     sandwich    & 6.33      \\
clock         & 4.5         &    pants       & 7.67    \\
apple         & 4.67        &    socks (sock) & 8.80      \\
tricycle      & 4.7         &   bowl        & 8.90       \\
 camera      & 4.78        &      &  \\
\bottomrule
\end{tabular}
}
\end{table}

\begin{table}[h]
\centering

\caption{\textbf{Out-of-Vocabulary Classes and Corresponding AoA Values.} The table lists the identified out-of-vocabulary classes along with their Age-of-Acquisition (AoA) values. For some classes, closely related words (shown in parentheses) were used to derive AoA values.}
\label{tab:ov-aoa} 
\resizebox{\linewidth}{!}{
\begin{tabular}{cccc}
\toprule
Vocab (Col 1) & AoA (Col 1) & Vocab (Col 2) & AoA (Col 2) \\
\midrule
sippycup (cup) & 3.57 &  collar & 6.56\\
toyrabbit (rabbit) & 3.94 & yarn & 6.61 \\
toyhorse (horse) & 4.15 & necktie & 6.63 \\
dresser & 4.28 & hanger & 6.78 \\
roadsign (sign) & 4.32 & binoculars & 6.79 \\
rug & 4.61 & telescope & 6.95 \\
doorknob & 4.70 & seashell & 7.06 \\
mask & 4.80 & golfball (golf) & 7.16 \\
dollhouse & 4.86 & dumbbell & 7.56 \\
muffins (muffin) & 5.11 & bathsuit (bathrobe) & 7.90 \\
tent & 5.16 & bowtie & 7.94 \\
hammer & 5.42 & rosary & 8.21 \\
frisbee & 5.50 & calculator & 8.22 \\
cushion & 5.53 & suitcase & 8.22 \\
watergun (gun) & 5.58 & trunk & 8.30 \\
ceilingfan (fan) & 5.63 & chessboard & 8.37 \\
helmet & 5.71 & compass & 8.44 \\
stapler & 5.83 & cupsaucer (saucer) & 8.44 \\
axe & 6.11 & lantern & 8.55 \\
speakers (speaker) & 6.11 & licenseplate (license) & 8.70 \\
lawnmower & 6.11 & pokercard (poker) & 9.10 \\
domino & 6.17 & keyboard & 9.32 \\
recordplayer & 6.37  & ringbinder (binder) & 10.42 \\
pitcher & 6.42  & powerstrip & 12.01 \\
 grill & 6.53  & & \\
\bottomrule
\end{tabular}
}
\end{table}

\subsection{Further Clarification on $n$-Way Classification Results} \label{sec:n-way_31_clarification}
In Figure~\ref{fig:in-out-of-vocab-nway}, the infant model CVCL (\inlinesymbol[0.3]{cvclgreen}{\linesolid}) using ``ours'' has limited class coverage. As shown in the Venn diagram (Figure~\ref{fig:venn_iv_oov}) and the coverage results (Figure~\ref{fig:class_coverage}), the ``in-vocab'' class in this setting is restricted. Consequently, the results include at most 31 classes. Therefore, in Figure~\ref{fig:in-out-of-vocab-nway}, the rightmost point for ``CVCL-ResNeXt50 (Ours)'' corresponds to $n=31$ instead of $n=32$.

\clearpage

\clearpage
\section{Centered Kernel Alignment (CKA)}\label{sec:cka-defined}
For two sets of activations, $\mathbf{X} \in \mathbb{R}^{n \times p}$ and $\mathbf{Y} \in \mathbb{R}^{n \times q}$, from corresponding layers of two models, where \( n \) is the number of examples and \( p \) and \( q \) are the feature dimensions (i.e., the number of neurons in each layer), the linear CKA is defined as: 
\begin{equation}
\operatorname{CKA}(\mathbf{X}, \mathbf{Y}) = \frac{\operatorname{HSIC}(\mathbf{X}, \mathbf{Y})}{\sqrt{\operatorname{HSIC}(\mathbf{X}, \mathbf{X}) \operatorname{HSIC}(\mathbf{Y}, \mathbf{Y})}},
\end{equation}
where $\operatorname{HSIC}(\cdot, \cdot)$ is the Hilbert-Schmidt Independence Criterion~\cite{gretton2005kernel}, which measures the dependence between two datasets. A higher CKA score indicates more similar representations between two models at the given layer.

\section{Call for More Openly Available Infant Dataset}\label{sec:callopendataset}

The success of infant computational models~\cite{vong2024grounded, orhan2020self, orhan2024learning} demonstrates the research potential of infant datasets like SAYCam~\cite{sullivan2021saycam}. While current access platform, \eg, Databrary\footnote{\url{https://databrary.org/}} (requiring institutional agreements) prioritize participant privacy, we identify an opportunity to expand access to inspire more research innovations. 

We call for more openly available infant datasets, similar to Ego4D~\cite{grauman2022ego4d} and Epic Kitchen~\cite{damen2018epickitchen}, while ensuring robust privacy safeguards (\eg, by blurring faces, removing other privacy-sensitive information, and muting any personally identifiable audio). Since modifying existing datasets for open access may be constrained by prior agreements, we encourage the development of new infant datasets with greater openness and sufficient privacy protection measures.

\end{appendices}

\end{document}